\documentclass[11pt,a4paper]{article}
\usepackage[hyperref]{emnlp2020}
\usepackage{times}
\usepackage{latexsym}

\usepackage{microtype}

\usepackage{booktabs}
\usepackage{multirow}
\usepackage{graphicx}
\usepackage{xcolor,colortbl}
\usepackage{enumitem}
\usepackage{amsmath}
\newcommand{\Mod}[1]{\ (\mathrm{mod}\ #1)}
\usepackage[ruled,noline]{algorithm2e}

\definecolor{Gray}{gray}{0.85}
\newcolumntype{a}{>{\columncolor{Gray}}c}

\aclfinalcopy 

\title{A Linguistic Analysis of Visually Grounded Dialogues \\
Based on Spatial Expressions}

\author{Takuma Udagawa$^1$ \: Takato Yamazaki$^{1}$ \: Akiko Aizawa$^{1,2}$ \\  
	The University of Tokyo, Tokyo, Japan$^1$ \\
	National Institute of Informatics, Tokyo, Japan$^2$\\
	\texttt{\{takuma\_udagawa,takatoy,aizawa\}@nii.ac.jp}}

\date{}

\begin{document}
\maketitle
\begin{abstract}
Recent models achieve promising results in visually grounded dialogues. However, existing datasets often contain undesirable biases and lack sophisticated linguistic analyses, which make it difficult to understand how well current models recognize their precise linguistic structures. To address this problem, we make two design choices: first, we focus on OneCommon Corpus \citep{udagawa2019natural,udagawa2020annotated}, a simple yet challenging common grounding dataset which contains minimal bias by design. Second, we analyze their linguistic structures based on \textit{spatial expressions} and provide comprehensive and reliable annotation for 600 dialogues. We show that our annotation captures important linguistic structures including predicate-argument structure, modification and ellipsis. In our experiments, we assess the model's understanding of these structures through reference resolution. We demonstrate that our annotation can reveal both the strengths and weaknesses of baseline models in essential levels of detail. Overall, we propose a novel framework and resource for investigating fine-grained language understanding in visually grounded dialogues.
\end{abstract}

\section{Introduction}
\label{sec:introduction}
\begin{figure*}[tb!]
\centering
\includegraphics[width=0.96\textwidth]{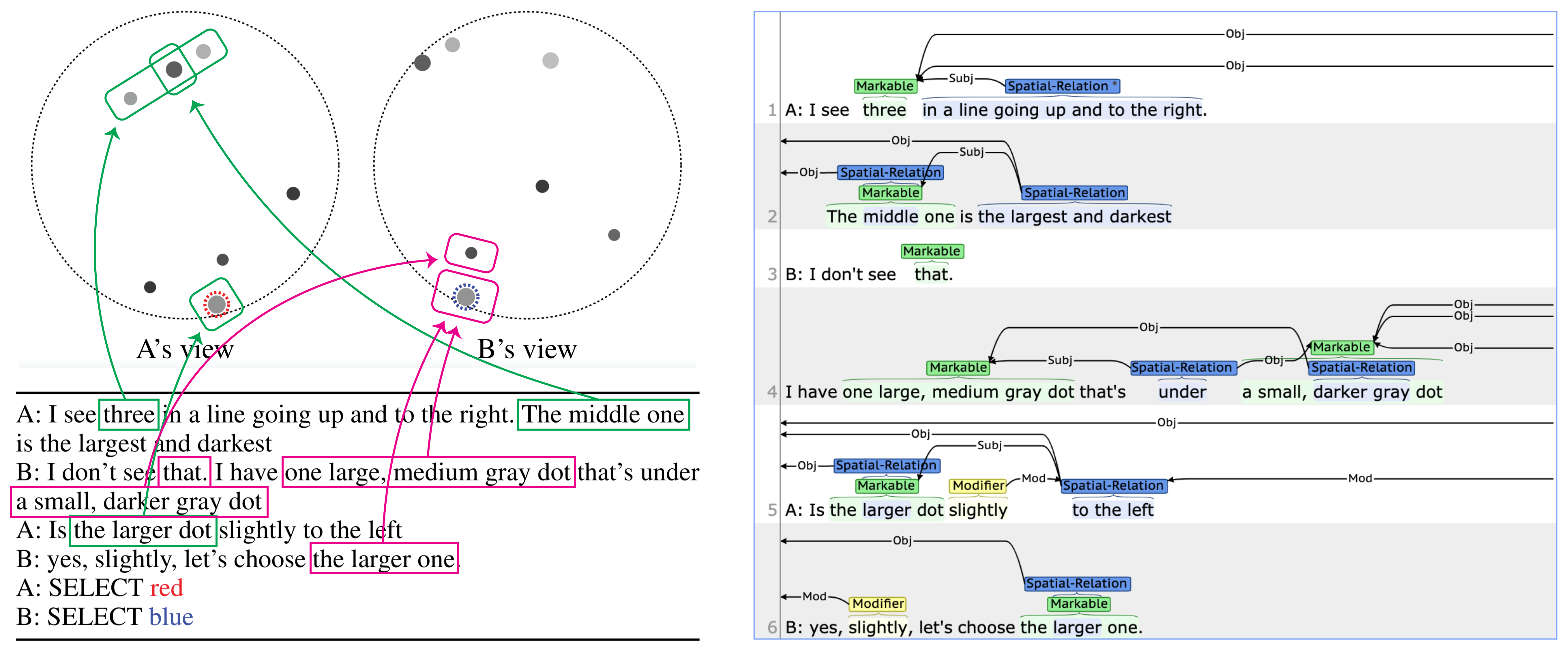}
\caption{Example dialogue from OneCommon Corpus with reference resolution annotation (left) and our spatial expression annotation (right). We consider spatial expressions as predicates and annotate their arguments as well as modifiers. For further details of the original dataset and our annotation schema, see Section \ref{sec:annotation}.
}
\label{fig:first_example}
\end{figure*}

Visual dialogue is the task of holding natural, often goal-oriented conversation in a visual context \citep{das2017visual,de2017guesswhat}. This typically involves two types of advanced grounding: \textit{symbol grounding} \citep{harnad1990symbol}, which bridges symbolic natural language and continuous visual perception, and \textit{common grounding} \citep{clark1996using}, which refers to the process of developing mutual understandings through successive dialogues. As noted in \citet{monroe2017colors,udagawa2019natural}, the \textit{continuous} nature of visual context introduces challenging symbol grounding of nuanced and pragmatic expressions. Some further incorporate \textit{partial observability} where the agents do not share the same context, which introduces complex misunderstandings that need to be resolved through advanced common grounding \citep{udagawa2019natural,haber-etal-2019-photobook}.

Despite the recent progress on these tasks, it remains unclear what types of linguistic structures can (or cannot) be properly recognized by existing models for two reasons. First, existing datasets often contain undesirable biases which make it possible to make correct predictions \textit{without} recognizing the precise linguistic structures \citep{goyal2017making,cirik-etal-2018-visual,agarwal-etal-2020-history}. Second, existing datasets severely lack in terms of sophisticated linguistic analysis, which makes it difficult to understand what types of linguistic structures exist or how they affect model performance.

To address this problem, we make the following design choices in this work:

\begin{itemize}[leftmargin=1em] \setlength{\parskip}{0pt}
\item We focus on OneCommon Corpus \citep{udagawa2019natural,udagawa2020annotated}, a simple yet challenging collaborative referring task under continuous and partially-observable context. In this dataset, the visual contexts are kept simple and controllable to remove undesirable biases while enhancing linguistic variety. In total, 5,191 natural dialogues are collected and fully annotated with referring expressions (which they called \textit{markables}) and their referents, which can be leveraged for further linguistic analysis.

\item To capture the linguistic structures in these dialogues, we propose to annotate \textit{spatial expressions} which play a central role in visually grounded dialogues. We take inspiration from the existing annotation frameworks \citep{pustejovsky2011iso,pustejovsky2011using,petruck-ellsworth-2018-representing,ulinski-etal-2019-spatialnet} but also make several simplifications and modifications to improve coverage, efficiency and reliability. \footnote{For instance, we define \textit{spatial expressions} in a broad sense and include spatial attributes (e.g. object size and color) as well as their comparisons.}
\end{itemize}

As shown in Figure \ref{fig:first_example}, we consider spatial expressions as \textit{predicates} with existing markables as their \textit{arguments}. We distinguish the argument roles based on \textit{subjects} and \textit{objects} \footnote{Our \textit{subject}-\textit{object} distinction corresponds to other terminologies such as \textit{trajector}-\textit{landmark} or \textit{figure}-\textit{ground}.} and annotate \textit{modifications} based on nuanced expressions (such as \textit{slightly}). By allowing the arguments to be in previous utterances, our annotation also captures \textit{argument ellipsis} in a natural way.

In our experiments, we focus on reference resolution to study the model's comprehension of these linguistic structures. Since we found the existing baseline to perform relatively poorly, we propose a simple method of incorporating \textit{numerical constraints} in model predictions, which significantly improved its prediction quality.

Based on our annotation, we conduct a series of analyses to investigate whether the model predictions are \textit{consistent} with the spatial expressions. Our main finding is that the model is adept at recognizing entity-level attributes (such as color and size), but mostly fails in capturing inter-entity relations (especially placements): using the terminologies from \citet{landau1993and}, the model can recognize the \textit{what} but not the \textit{where} in spatial language. We also conduct further analyses to investigate the effect of other linguistic factors.

Overall, we propose a novel framework and resource for conducting fine-grained linguistic analyses in visually grounded dialogues. All materials in this work will be publicly available at \url{https://github.com/Alab-NII/onecommon} to facilitate future model development and analyses.

\section{Related Work}
\label{sec:related_work}

Linguistic structure plays a critical role in dialogue research. From theoretical aspects, various dialogue structures have been studied, including discourse structure \citep{stent-2000-rhetorical,asher2003logics}, speech act \citep{austin1975things,searle1969speech} and common grounding \citep{clark1996using,lascarides2009agreement}. In dialogue system engineering, various linguistic structures have been considered and applied, including syntactic dependency \citep{davidson-etal-2019-dependency}, predicate-argument structure (PAS) \citep{yoshino2011spoken}, ellipsis \citep{quan-etal-2019-gecor,hansen2020you}, intent recognition \citep{silva2011symbolic,shi-etal-2016-deep}, semantic representation/parsing \citep{mesnil2013investigation,gupta-etal-2018-semantic} and frame-based dialogue state tracking \citep{williams2016dialog,elasri2017frames}. However, most prior work focus on dialogues where information is not grounded in external, perceptual modality such as vision. In this work, we propose an effective method of analyzing linguistic structures in visually grounded dialogues.

Recent years have witnessed an increasing attention in visually grounded dialogues \citep{zarriess-etal-2016-pentoref,de2018talk,alamri2019audio,narayan-chen-etal-2019-collaborative}. Despite the impressive progress on benchmark scores and model architectures \citep{Das_2017_ICCV,Wu_2018_CVPR,Kottur_2018_ECCV,gan-etal-2019-multi,shukla-etal-2019-ask,niu2019recursive,zheng2019reasoning,kang-etal-2019-dual,visdial_bert,pang2020visual}, there have also been critical problems pointed out in terms of dataset biases \citep{goyal2017making,visdial_eval,massiceti2018visual,chen-etal-2018-attacking,kottur-etal-2019-clevr,kim2020modality,agarwal-etal-2020-history} which obscure such contributions. For instance, \citet{cirik-etal-2018-visual} points out that existing dataset of reference resolution may be largely solvable \textit{without} recognizing the full referring expressions (e.g. based on object categories only). To circumvent these issues, we focus on OneCommon Corpus where the visual contents are simple (exploitable categories are removed) and well-balanced (by sampling from uniform distributions) to minimize dataset biases.

Although various probing methods have been proposed for models and datasets in NLP \citep{belinkov:2019:tacl,geva-etal-2019-modeling,kaushik2019learning,gardner2020evaluating,ribeiro-etal-2020-beyond}, fine-grained analyses of visually grounded dialogues have been relatively limited. Instead, \citet{kottur-etal-2019-clevr} proposed a diagnostic dataset to investigate model's language understanding: however, their dialogues are generated artificially and may not reflect the true nature of visual dialogues. \citet{shekhar-etal-2019-beyond} also acknowledges the importance of linguistic analysis but only dealt with coarse-level features that can be computed automatically (such as dialogue topic and diversity). Most similar and related to our research are \citet{yu-etal-2019-see,udagawa2020annotated}, where they conducted additional annotation of reference resolution in visual dialogues: however, they still do not capture more sophisticated linguistic structures such as PAS, modification and ellipsis.

Finally, spatial language and cognition have a long history of research \citep{talmy1983language,herskovits1987language}. In computational linguistics, \citep{kordjamshidi-etal-2010-spatial,pustejovsky2015semeval} developed the task of spatial role labeling to capture spatial information in text: however, they do not fully address the problem of annotation reliability nor grounding in external visual modality. In computer vision, the VisualGenome dataset \citep{krishna2017visual} provides rich annotation of spatial scene graphs constructed from raw images, but not from raw dialogues. \citet{ramisa-etal-2015-combining,platonov2018computational} also worked on modelling spatial prepositions in single sentences. To the best of our knowledge, our work is the first to apply, model and analyze spatial expressions in visually grounded dialogues at full scale.

\section{Annotation}
\label{sec:annotation}

\subsection{Dataset}
\label{subsec:dataset}

Our work extends OneCommon Corpus originally proposed in \citet{udagawa2019natural}. In this task, two players A and B are given slightly different, overlapping perspectives of a 2-dimensional grid with 7 entities in each view (Figure \ref{fig:first_example}, left). Since only some (4, 5 or 6) of them are in common, this setting is \textit{partially-observable} where complex misunderstandings and partial understandings are introduced. In addition, each entity only has \textit{continuous} attributes (x-value, y-value, color and size), which introduce various nuanced and pragmatic expressions. Note that all entity attributes are generated randomly to enhance linguistic diversity and reduce dataset biases. Under this setting, two players were instructed to converse freely in natural language to coordinate attention on one of the same, common entities. Basic statistics of the dialogues are shown at the top of Table \ref{onecommon_statistics} and the frequency of nuanced expressions estimated in Table \ref{nuanced_statistics}.

\begin{table}[th]
\centering \small
\begin{tabular}{cc}
\toprule
Total dialogues & 6,760 \\
Avg. utterances per dialogue & 4.76 \\
Avg. tokens per utterance & 12.37 \\
\midrule
Successful dialogues & 5,191 \\
Annotated markables & 40,172 \\
\% markables with 1 referent\phantom{s$>$} & 71.81 \\
\% markables with 2 referents\phantom{$>$} & 14.85 \\
\% markables with $\geq$3 referents & 12.03 \\
\% markables with 0 referent\phantom{s$>$} & 1.31 \\
\bottomrule
\end{tabular}
\caption{
OneCommon Corpus statistics.
}
\label{onecommon_statistics}
\end{table}

\begin{table}[ht]
\centering \small
\begin{tabular}{lcc}
\toprule
Nuance Type & \% Utterance & Example Usage \\
\midrule
Approximation & 3.98 & \textbf{almost} in the middle \\
Exactness & 2.71 & \textbf{exactly} horizontal \\
Subtlety & 9.37 & \textbf{slightly} to the right \\
Extremity & 9.35 & \textbf{very} light dot \\
Uncertainty & 5.79 & \textbf{Maybe} it's different \\
\bottomrule
\end{tabular}
\caption{\label{nuanced_statistics}
Estimated frequency of nuanced expressions from \citet{udagawa2019natural}.
}
\end{table}

More recently, \citet{udagawa2020annotated} curated all successful dialogues from the corpus and additionally conducted reference resolution annotation. Specifically, they detected all referring expressions (\textit{markables}) based on minimal noun phrases by trained annotators and identified their referents by multiple crowdworkers (Figure \ref{fig:first_example} left, highlighted). Both annotations were shown to be reliable with high overall agreement. We show their dataset statistics at the bottom of Table \ref{onecommon_statistics}.

In this work, we randomly sample 600 dialogues from the latest corpus (5,191 dialogues annotated with reference resolution) to conduct further annotation of spatial expressions.

\subsection{Annotation Schema}
\label{subsec:annotation_schema}

Our annotation procedure consists of three steps: \textit{spatial expression detection}, \textit{argument identification} and \textit{canonicalization}. Based on these annotation, we conduct fine-grained analyses of the dataset (Subsection \ref{subsec:annotation_results}) as well as the baseline models (Subsection \ref{subsec:model_analysis}). For further details and examples of our annotation, see Appendix \ref{sec:annotation_examples}.

\subsubsection{Spatial Expression Detection}
\label{subsubsec:spatial_expression_detection}

Based on the definition from \citet{pustejovsky2011iso,pustejovsky2011using}, spatial expressions are ``constructions that make explicit reference to the spatial attributes of an object or spatial relations between objects''. \footnote{Note that their term \textit{object} corresponds to our term \textit{entity}.} We generally follow this definition and detect all spans of spatial attributes and relations in the dialogue. To make the distinction clear, we consider entity-level information like color and size as spatial attributes, and other information such as location and \textit{explicit} attribute comparison as spatial relations. Spatial attributes could be annotated as adjectives (``\textit{dark}''), prepositional phrases (``\textit{of light color}'') or noun phrases (``\textit{a black dot}''), while spatial relations could be adjectives (``\textit{lighter}''), prepositions (``\textit{near}''), and so on. We also detect modifiers of spatial expressions based on nuanced expressions (c.f. Table \ref{nuanced_statistics}).

Although we allow certain flexibility in determining their spans, holistic/dependent expressions (such as ``\textit{all shades of gray}'', ``\textit{sloping up to the right}'', ``\textit{very slightly}'') were instructed to be annotated as a single span. Independent expressions (e.g. connected by conjunctions) could be annotated separately or jointly if they had the same structure (e.g. same arguments and modifiers).

For the sake of efficiency, we \underline{do not} annotate spatial attributes and their modifiers inside markables (see Figure \ref{fig:first_example}), since their spans and arguments are easy to be detected automatically.

\subsubsection{Argument Identification}
\label{subsubsec:argument_identification}

Secondly, we consider the detected spatial expressions as \textit{predicates} and annotate referring expressions (markables) as their \textit{arguments}. This approach has several advantages: first, it has broad coverage since referring expressions are prevalent in visual dialogues. In addition, by leveraging \textit{exophoric} references which directly bridge natural language and the visual context, we can conduct essential analyses related to symbol grounding across the two modalities (Subsection \ref{subsec:model_analysis}).

To be specific, we distinguish the argument roles based on subjects and objects. We allow arguments to be in previous utterances \textit{only if} they are unavailable in the present utterance. Multiple markables can be annotated for the subject/object roles, and no object need to be annotated in cases of spatial attributes, nominal/verbal expressions (``\textit{triangle}'', ``\textit{clustered}'') or \textit{implicit global objects} as in superlatives (``\textit{darkest} (of all)''). If the arguments are indeterminable based on these roles (as in enumeration, e.g. ``\textit{\underline{From left to right}, there are ...}''), they were marked as \textit{unannotatable}. Modificands of the modifiers (which could be either spatial attributes or relations) were also identified in this step.

\subsubsection{Canonicalization}
\label{subsubsec:canonicalization}

Finally, we conduct canonicalization of the spatial expressions and modifiers. Since developing a complete ontology for this domain is infeasible or too expensive, we focus on canonicalizing the central \textit{spatial relations} in this work: we \underline{do not} canonicalize spatial attributes manually, since we can canonicalize the central spatial attributes automatically (c.f. Subsubsection \ref{subsubsec:spatial_attributes}).

According to \citet{landau2017update}, there are 2 classes of relations in spatial language: \textit{functional} class whose core meanings engage force-dynamic relationship (such as \textit{on}, \textit{in}) and \textit{geometric} class whose core meanings engage geometry (such as \textit{left}, \textit{above}). Since functional relations are less common in this dataset and more difficult to define due to their vagueness and context dependence \citep{platonov2018computational}, we focus on the following 5 categories of geometric relations and attribute comparisons, including a total of 24 canonical relations which can be defined explicitly.

\textbf{Direction} requires the subjects and objects to be placed in certain orientation: \textit{left}, \textit{right}, \textit{above}, \textit{below}, \textit{horizontal}, \textit{vertical}, \textit{diagonal}.

\textbf{Proximity} is related to distance between subjects, objects or other entities: \textit{near}, \textit{far}, \textit{alone}.

\textbf{Region} restricts the subjects to be in a certain region specified by the objects: \textit{interior}, \textit{exterior}.

\textbf{Color comparison} is related to comparison of color between subjects and objects: \textit{lighter}, \textit{lightest}, \textit{darker}, \textit{darkest}, \textit{same color}, \textit{different color}.

\textbf{Size comparison} is related to comparison of size between subjects and objects: \textit{smaller}, \textit{smallest}, \textit{larger}, \textit{largest}, \textit{same size}, \textit{different size}.

To be specific, we annotate whether each detected spatial relation \textit{implies} any of the 24 canonical relations. Each spatial relation can imply multiple canonical relations (e.g. ``on the upper right'' implies \textit{right} and \textit{above}) or none (e.g. ``triangle'' does not imply any of the above relations).

In addition, we define 6 modification types (\textit{subtlety}, \textit{extremity}, \textit{uncertainty}, \textit{certainty}, \textit{neutrality} and \textit{negation}) and canonicalize each modifier into one type. For example, ``very slightly'' is considered to have the overall type of \textit{subtlety}.

\subsection{Results}
\label{subsec:annotation_results}

\subsubsection{Annotation Reliability}
\label{subsubsec:annotation_reliability}

\begin{table}[ht]
\centering \small
\begin{tabular}{lcc}
\toprule
Annotation & \% Agreement & Cohen's $\kappa$ \\
 \midrule
Attribute Span & 98.5 & 0.88 \\
Relation Span & 95.1 & 0.87 \\
Modifier Span & 99.2 & 0.86 \\
\midrule
Subject Ident. & 98.8 & 0.96 \\
Object Ident. & 95.9 & 0.79 \\
Modificand Ident. & 99.6 & 0.98 \\
\midrule
Relation Canon. & 99.7 & 0.96 \\
Modifier Canon. & 87.5 & 0.83 \\
\bottomrule
\end{tabular}
\caption{
Results of our reliability analysis.
}
\label{reliability_results}
\end{table}

To test the reliability of our annotation, two trained annotators (the authors) independently detected the spatial expressions and modifiers in 50 dialogues. Then, using the 50 dialogues from one of the annotators, two annotators independently conducted argument identification and canonicalization. We show the observed agreement and Cohen's $\kappa$ \cite{cohen1968weighted} in Table \ref{reliability_results}.

For span detection, we computed the token level agreement of spatial expressions and modifiers. Despite having certain freedom for determining their spans, we observed very high agreement (including their starting positions, see Appendix \ref{sec:annotation_results}).

For argument identification, we computed the exact match rate of the arguments and modificands. As a result, we observed near perfect agreement for subject/modificand identification. For object identification, the result seems relatively worse: however, upon further inspection, we verified that 73.5\% of the disagreements were essentially based on the same markables (e.g. coreferences).

Finally, we observed reasonably high agreement for relation/modifier canonicalization as well. Overall, we conclude that all steps of our annotation can be conducted with high reliability.

\subsubsection{Annotation Statistics}
\label{subsubsec:annotation_statistics}

\begin{table}[th]
\centering \small
\begin{tabular}{ccc}
\toprule
 & Attribute & Relation \\
 \midrule
Total & 378 & 4,300 \\
Unique & 121 & 1,139 \\
Avg. per dialogue & 0.63 & 7.17 \\
\% inter-utterance subject & 1.59 & 1.37 \\
\% inter-utterance object & - & 14.65 \\
\% no object & - & 30.84 \\
\% modified & 36.51 & 16.86 \\
\% unannotatable & 0.79 & 0.79 \\
\bottomrule
\end{tabular}
\caption{
Statistics of our spatial expression annotation in 600 randomly sampled dialogues.
}
\label{annotation_statistics}
\end{table}

The basic statistics of our annotation are summarized in Table \ref{annotation_statistics}. Note that there are relatively few spatial attributes annotated, since most of them appeared inside the markables (hence not detected manually). However, a large number of spatial relations with non-obvious structures were identified.

In both expressions, we found over 1\% of the subjects and 14\% of the objects to be present only in previous utterances, which indicates that argument level ellipses are common and need to be resolved in visual dialogues. For spatial relations, about 30\% did not have any explicit objects.

Our annotation also verified that a large portion of the spatial expressions (37\% for spatial attributes and 17\% for relations) accompanied modifiers. 

Finally, less than 1\% of spatial expressions were \textit{unannotatable} based on our schema, which verifies its broad coverage. Overall, our annotation can capture important linguistic structures of visually grounded dialogues, and it is straightforward to conduct even further analyses (e.g. by focusing on specific canonical relations or modifications).

\section{Experiments}
\label{sec:experiments}

\subsection{Reference Resolution}
\label{subsec:reference_resolution}

Reference resolution is an important subtask of visual dialogue that can be used for probing model's understanding of intermediate dialogue process \citep{udagawa2020annotated}. As illustrated in Figure \ref{fig:first_example} (left), this is a simple task of predicting the referents for each markable based on the \textit{speaker}'s perspective. To collect model predictions for all dialogues, we split the whole dataset into 10 equal-sized bins and use each bin as the test set in 10 rounds of the experiments. For a more detailed setup of our experiments, see Appendix \ref{sec:experiment_setup}.

\subsubsection{Models}
\label{subsubsec:models}

\begin{figure}[ht]
\centering
\includegraphics[width=0.98\columnwidth]{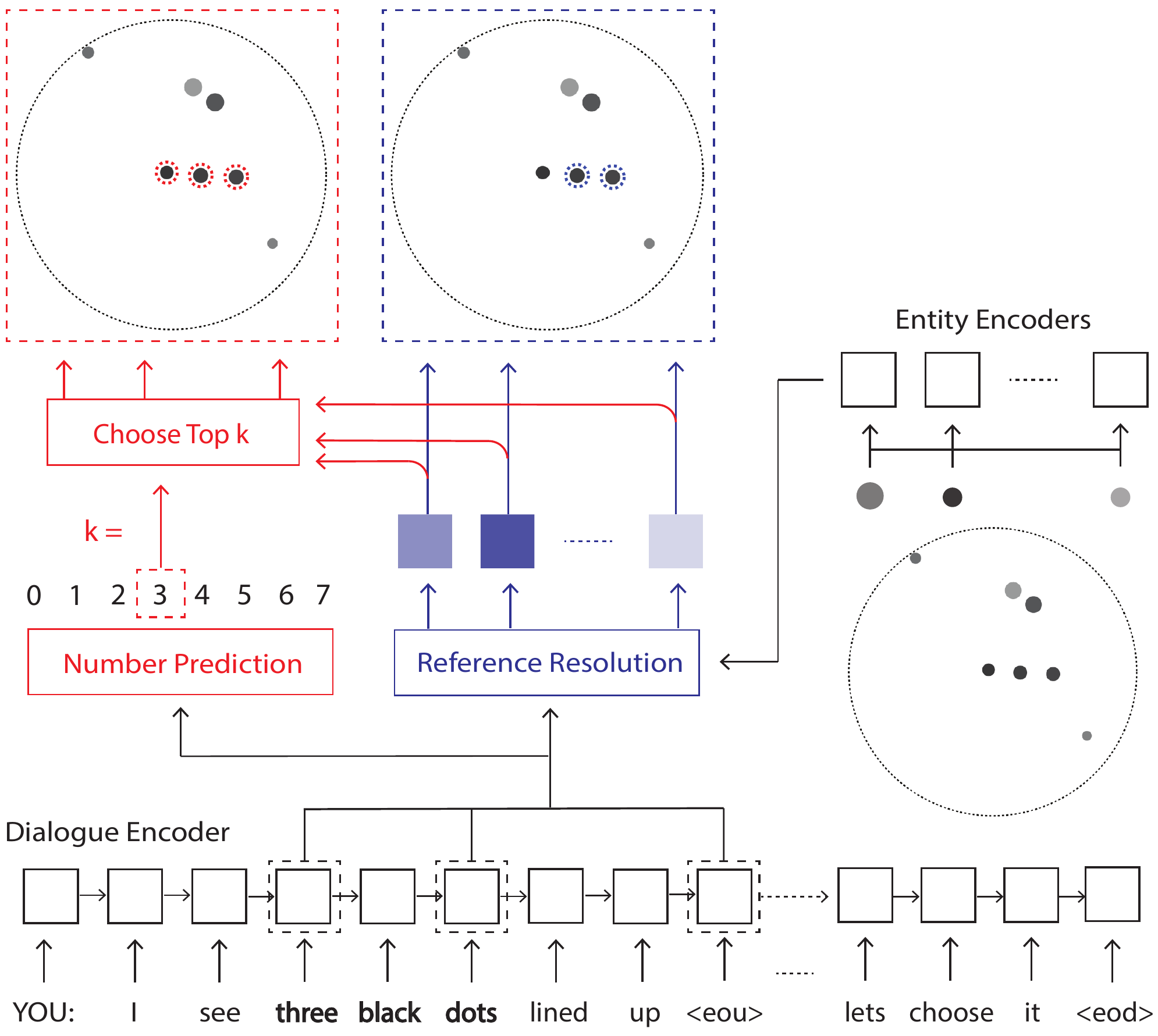}
\caption{Our model architecture. \texttt{REF} prediction flow is shown in blue and \texttt{NUMREF} prediction flow in red.}
\label{fig:model_architecture}
\end{figure}

As a baseline, we use the \texttt{REF} model proposed in \citet{udagawa2020annotated}. As shown in Figure \ref{fig:model_architecture}, this model has two encoders: \textit{dialogue encoder} based on a simple GRU \citep{cho2014properties} and \textit{entity encoder} which outputs entity-level representation of the observation based on MLP and relational network \citep{santoro2017simple}. To predict the referents, \texttt{REF} takes the GRU's start position of the markable, end position of the markable and end position of the utterance to compute entity-level scores and judge whether each entity is a referent based on logistic regression.

However, since the predictions are made independently for each entity, this model often predicts the wrong number of referents, leading to low performance in terms of exact match rate. To address this issue, we trained a separate module to track the \textit{number} of referents in each markable. We formulate this as a simple classification task between 0, 1, ..., 7, which can be predicted reliably with an average accuracy of $92$\%. Based on this module's prediction $k$, we simply take the top $k$ entities with the highest scores as the referents. We refer to this numerically constrained model as \texttt{NUMREF}.

Furthermore, we conduct feature level ablations to study the importance of each feature: for instance, we remove the xy-values from the structured input to ablate the \textit{location} feature.

\subsubsection{Results}
\label{subsubsec:reference_resolution_results}

\begin{table}[ht]
\centering \small
\begin{tabular}{lcc}
\toprule
 & Entity-Level & Markable-Level \\
 & Accuracy & Exact Match \\
\midrule
\texttt{REF} & 85.71$\pm$0.23 & 33.15$\pm$1.00 \\
{\texttt{REF}$-$location} & 84.28$\pm$0.27 & 30.53$\pm$0.84 \\
{\texttt{REF}$-$color} & 83.08$\pm$0.32 & 17.09$\pm$1.04 \\
{\texttt{REF}$-$size} & 83.50$\pm$0.22 & 19.41$\pm$0.98 \\
\midrule
\texttt{NUMREF} & \textbf{86.03$\pm$0.33} & \textbf{54.94$\pm$0.76} \\
{\texttt{NUMREF}$-$location} & 83.35$\pm$0.26 & 49.77$\pm$0.64 \\
{\texttt{NUMREF}$-$color} & 81.19$\pm$0.41 & 39.74$\pm$1.31 \\
{\texttt{NUMREF}$-$size} & 82.39$\pm$0.20 & 43.40$\pm$0.67 \\
\midrule
Human & 96.26 & 86.90 \\
\bottomrule
\end{tabular}
\caption{
Reference resolution results.
}
\label{experiment_results}
\end{table}

We report the mean and standard deviation of the entity-level accuracy and markable-level exact match rate in Table \ref{experiment_results}. Compared to \texttt{REF}, our \texttt{NUMREF} model slightly improves the entity-level accuracy and significantly outperforms it in terms of exact match rate, which validates our motivation. From the ablation studies, we can see that all features contribute to the overall performance, but color and size seem to have the largest impact. 

However, it is difficult to see how and where these models struggle based on mere accuracy. For further investigation, we need more sophisticated \textit{behavioral testing} (namely black-box testing) to verify whether each model has the capability of recognizing certain concepts or linguistic structures \citep{ribeiro-etal-2020-beyond}.

\subsection{Model Analysis}
\label{subsec:model_analysis}

To study the current model's strengths and weaknesses in detail, we investigate whether their predictions are \textit{consistent} with the central spatial expressions.

\subsubsection{Spatial Attributes}
\label{subsubsec:spatial_attributes}

\begin{figure}[ht]
\centering
\includegraphics[width=0.96\columnwidth]{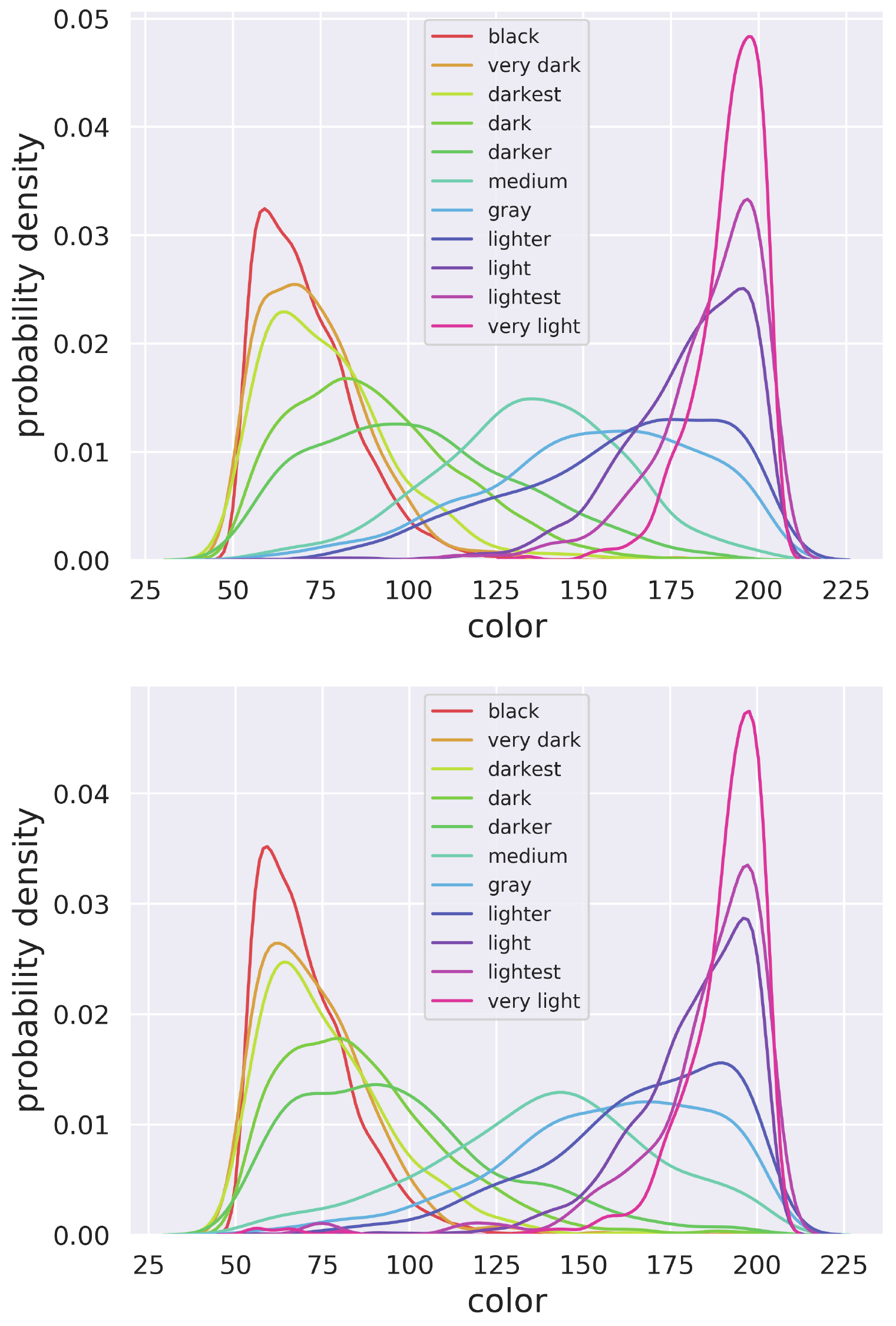}
\caption{Referent color distributions. Top is human, bottom is \texttt{NUMREF} (smaller is darker in color axis).}
\label{fig:referent_color}
\end{figure}

First, we analyze whether the model predictions are consistent with the entity-level spatial attributes. Since most of them were confirmed to appear inside the markables (Subsection \ref{subsec:annotation_results}), we automatically detect all expressions of \textit{color} in the markables, plot the distributions of the actual referent color, and compare the results between gold human annotation and model predictions (Figure \ref{fig:referent_color}).

From the figure, we can verify that the two distributions look almost identical for the common color expressions, and our \texttt{NUMREF} model seems to capture important characteristics of pragmatic expressions (same expression being used for wide range of colors) and modifications such as neutrality (\textit{medium}) and extremity (\textit{very dark}, \textit{very light}). \footnote{Spatial attributes with nuances of subtlety (such as \textit{slightly dark}) were relatively rare and omitted in the figure.} We observed very similar result with the \textit{size} distributions, which is available in Appendix \ref{sec:size_distribution_plots}.

Based on these results, we argue that the current model can capture entity-level attributes very well, including basic modification.

\subsubsection{Spatial Relations}
\label{subsubsec:spatial_relations}

\begin{table*}[th]
\centering \small
\def\arraystretch{1.0}
\newcommand{\intermidrule}{\cmidrule{2-13}}  
\setlength{\tabcolsep}{4pt}
\setlength{\aboverulesep}{0pt}
\setlength{\belowrulesep}{0pt}
\setlength{\extrarowheight}{.75ex}
\begin{tabular}{lcc|cacacacaca}
\toprule
\multicolumn{3}{c}{Models} & \multicolumn{2}{c}{\texttt{REF}} & \multicolumn{2}{c}{\texttt{REF}-abl} & \multicolumn{2}{c}{\texttt{NUMREF}} & \multicolumn{2}{c}{\texttt{NUMREF}-abl} & \multicolumn{2}{c}{Human} \\
\midrule
Category & Relation & \# Cases  & satisfy & \phantom{.}valid\phantom{.} & satisfy & \phantom{.}valid\phantom{.} & satisfy & \phantom{.}valid\phantom{.} & satisfy & \phantom{.}valid\phantom{.} & satisfy & \phantom{.}valid\phantom{.} \\
\midrule
\multirow{8}{*}{Direction}\phantom{..} & \textit{left} & 412 & 23.5 & 32.3 & 21.1 & 28.9 & \textbf{67.0} & \textbf{99.5} & 62.4 & \textbf{99.5} & 95.9 & 97.6 \\
 & \textit{right} & 468 & 28.0 & 35.5 & 24.6 & 30.8 & 67.3 & \textbf{98.7} & \textbf{68.2} & \textbf{98.7} & 95.3 & 96.4 \\
 & \textit{above} & 514 & 28.6 & 37.4 & 24.7 & 33.1 & 65.2 & 99.2 & \textbf{66.5} & \textbf{99.4} & 96.7 & 98.6 \\
 & \textit{below} & 444 & 25.2 & 34.5 & 21.6 & 27.9 & \textbf{66.0} & \textbf{99.1} & 62.2 & \textbf{99.1} & 96.4 & 96.8 \\
 & \textit{horizontal} & 37 & 54.1 & 70.3 & 27.0 & 59.5 & \textbf{59.5} & \textbf{100.0} & 51.4 & 97.3 & 91.9 & 100.0 \\
 & \textit{vertical} & 46 & 37.0 & 73.9 & 23.9 & 54.3 & 43.5 & \textbf{95.7} & \textbf{45.7} & \textbf{95.7} & 82.6 & 100.0 \\
 & \textit{diagonal} & 50 & 48.0 & 74.0 & 30.0 & 50.0 & \textbf{60.0} & \textbf{98.0} & \textbf{60.0} & \textbf{98.0} & 90.0 & 100.0 \\
\intermidrule
 & All & 1,971 & 27.8 & 37.6 & 23.4 & 31.9 & \textbf{65.5} & \textbf{99.0} & 64.1 & \textbf{99.0} & 95.5 & 97.6 \\
\midrule
\multirow{4}{*}{Proximity} & \textit{near} & 271 & 49.4 & 61.3 & 29.9 & 49.1 & \textbf{77.1} & 94.5 & 56.1 & \textbf{95.2} & 95.2 & 96.7 \\
 & \textit{far} & 27 & 29.6 & 40.7 & 33.3 & 40.7 & 77.8 & \textbf{100.0} & \textbf{92.6} & \textbf{100.0} & 96.3 & 96.3 \\
 & \textit{alone} & 111 & 36.9 & 44.1 & 45.0 & 54.1 & \textbf{68.5} & \textbf{94.6} & 67.6 &\textbf{94.6} & 91.9 & 94.6 \\
\intermidrule
 & All & 409 & 44.7 & 55.3 & 34.2 & 49.9 & \textbf{74.8} & 94.9 & 61.6 & \textbf{95.4} & 94.4 & 96.1 \\
\midrule
\multirow{3}{*}{Region} & \textit{interior} & 135 & 38.5 & 52.6 & 27.4 & 39.3 & \textbf{62.2} & 93.3 & 58.5 & \textbf{94.1} & 96.3 & 100.0 \\
 & \textit{exterior} & 62 & 40.3 & 48.4 & 40.3 & 53.2 & 80.6 & \textbf{98.4} & \textbf{87.1} & \textbf{98.4} & 98.4 & 98.4 \\
\intermidrule
 & All & 197 & 39.1 & 51.3 & 31.5 & 43.7 & \textbf{68.0} & 94.9 & 67.5 & \textbf{95.4} & 97.0 & 99.5 \\
\midrule
\multirow{7}{*}{Color} & \textit{lighter} & 147 & 23.1 & 25.9 & 6.8 & 8.2 & \textbf{84.4} & \textbf{100.0} & 57.1 & 99.3 & 97.3 & 98.0 \\
 & \textit{lightest} & 42 & 45.2 & 66.7 & 14.3 & 33.3 & \textbf{61.9} & \textbf{100.0} & 31.0 & \textbf{100.0} & 83.3 & 100.0 \\
 & \textit{darker} & 171 & 24.0 & 26.3 & 7.0 & 10.5 & \textbf{83.0} & \textbf{99.4} & 53.2 & \textbf{99.4} & 95.9 & 98.8 \\
 & \textit{darkest} & 48 & 56.2 & 64.6 & 14.6 & 33.3 & \textbf{66.7} & \textbf{100.0} & 35.4 & \textbf{100.0} & 89.6 & 97.9 \\
 & \textit{same} & 50 & 12.0 & 30.0 & 8.0 & 30.0 & \textbf{40.0} & \textbf{88.0} & 32.0 & 86.0 & 92.0 & 96.0 \\
 & \textit{different} & 14 & 64.3 & 71.4 & 71.4 & 71.4 & 64.3 & \textbf{100.0} & \textbf{78.6} & 92.9 & 92.9 & 100.0 \\
\intermidrule
 & All & 472 & 28.8 & 35.4 & 10.4 & 18.0 & \textbf{74.8} & \textbf{98.5} & 49.2 & 97.9 & 94.1 & 98.3 \\
\midrule
\multirow{7}{*}{Size} & \textit{smaller} & 213 & 27.7 & 31.5 & 7.5 & 9.9 & \textbf{80.8} & \textbf{100.0} & 59.6 & \textbf{100.0} & 98.6 & 99.5 \\
 & \textit{smallest} & 52 & 71.2 & 73.1 & 21.2 & 34.6 & \textbf{86.5} & \textbf{98.1} & 48.1 & \textbf{98.1} & 92.3 & 98.1 \\
 & \textit{larger} & 238 & 23.1 & 28.6 & 9.7 & 16.0 & \textbf{73.5} & \textbf{99.6} & 48.7 & \textbf{99.6} & 98.3 & 98.3 \\
 & \textit{largest} & 61 & 52.5 & 60.7 & 11.5 & 24.6 & \textbf{73.8} & \textbf{100.0} & 39.3 & \textbf{100.0} & 96.7 & 100.0 \\
 & \textit{same} & 103 & 34.0 & 42.7 & 18.4 & 27.2 & \textbf{80.6} & 88.3 & 65.0 & \textbf{91.3} & 98.1 & 100.0 \\
 & \textit{different} & 12 & 75.0 & 75.0 & 66.7 & 66.7 & \textbf{91.7} & \textbf{91.7} & 83.3 & 83.3 & 91.7 & 91.7 \\
\intermidrule
 & All & 679 & 33.4 & 38.7 & 12.4 & 18.9 & \textbf{78.2} & 97.8 & 54.3 & \textbf{98.1} & 97.6 & 99.0 \\
\bottomrule
\end{tabular}
\caption{
Canonical relation test results. We compute the \textit{satisfy} and \textit{valid} rate of the predictions for each canonical relation. Best scores of the models are in bold (-abl shows the corresponding feature ablated results).
}
\label{satisfication_result}
\end{table*}

Next, we investigate whether the model predictions are consistent with the central spatial relations. Based on our annotation (Subsection \ref{subsec:annotation_schema}), we conduct simple tests to check whether the predicted referents satisfy each canonical relation. To be specific, our tests check for two conditions: whether the predictions are \textit{valid} (satisfy the minimal requirements, e.g. at least 2 referents predicted for \textit{near} relation), and if they are valid, whether the predictions actually \textit{satisfy} the canonical relation (e.g. referents are closer than a certain threshold).

Algorithm \ref{alg:left} shows our test for the canonical \textit{left} relation. Note that if no objects are annotated, we simply test whether the referents are on the left side of the player's view. For further details/examples of our canonical relation tests, see Appendix \ref{sec:relation_test_algorithms}.

\begin{algorithm}[h]
\small
\DontPrintSemicolon
\SetAlgoNoEnd

\KwIn{subject referents $\mathcal{S}$, object referents $\mathcal{O}$, boolean $no\_object$}
\KwOut{boolean $satisfy$, boolean $valid$}
\If{$no\_object$}{
	$valid \leftarrow |\mathcal{S}| \! > \! 0$\\
	$satisfy \leftarrow valid \, \wedge \, mean(\mathcal{S}.x) \! < \! 0$
}\Else{
	$valid \leftarrow |\mathcal{S}| \! > \! 0 \, \wedge \, |\mathcal{O}| \! > \! 0$\\
	$satisfy \leftarrow valid \, \wedge \, mean(\mathcal{S}.x) \! < \! mean(\mathcal{O}.x)$
}
\Return $satisfy$, $valid$
\caption{Test for \textit{left} relation}
\label{alg:left}
\end{algorithm}

The results of our tests are summarized in Table \ref{satisfication_result}. We also compare with the feature ablated models to estimate the test cases which can be satisfied \textit{without} using the corresponding features, i.e. location for \textit{direction}/\textit{proximity}/\textit{region} categories, color for \textit{color comparison}, and size for \textit{size comparison}.

First, we can verify that human annotation passes most of our tests, which is an important evidence of the \textit{validity} of our annotations and tests. We also confirmed that \texttt{REF} models often make \textit{invalid} predictions with overall poor performance, which is consistent with our expectation.

In \textit{direction}, \textit{proximity} and \textit{region} categories, we found that \texttt{NUMREF} model performs on par or only marginally better than its ablated version (and even underperforms it for simple relations like \textit{right} and \textit{above}): these results indicate that current model is still incapable of leveraging locational features to make more consistent predictions.  \footnote{For relations like \textit{far} and \textit{different color}, ablated model may be better simply because referents tend to be more distant/dissimilar when predictions are closer to random.}

In \textit{color/size comparison}, \texttt{NUMREF} performs reasonably well, outperforming all other models: this indicates that the model can not only capture but also \textit{compare} entity-level attributes to a certain extent. However, there is still room left for improvement in almost all relations. It is also worth noting that \textit{size comparison} may be easier, as the range of size is limited (only \textit{6} compared to \textit{150} for color).

Overall, we conclude that current models still struggle in capturing most of the inter-entity relations, especially those related to placements.

\subsubsection{Further Analyses}
\label{subsubsec:further_analyses}

\begin{table}[h!]
\centering \small
\begin{tabular}{lccc}
\toprule
Linguistic Factors & \# Cases & \texttt{NUMREF} & Human  \\
 \midrule
strong modification & 149 & 76.51 & 95.97 \\
neutral & 3,094 & 70.46 & 95.77 \\
weak modification & 490 & 66.12 & 95.10 \\
\midrule
inter-utterance subject & 14 & 57.14 & 92.86 \\
inter-utterance object & 265 & 72.08 & 94.72 \\
no object & 1,127 & 74.45 & 92.99 \\
ignorable object & 1,805 & 69.64 & 97.23 \\
unignorable object & 796 & 65.33 & 96.11 \\
\midrule
All & 3,728 & 70.17 & 95.71 \\
\bottomrule
\end{tabular}
\caption{
Satisfy rate classified by linguistic factors.
}
\label{further comparison}
\end{table}

Finally, we conduct further analyses to study other linguistic factors that affect model performance. Table \ref{further comparison} shows the results of our relation tests classified by notable linguistic structures.

In terms of modification, we can confirm that human performance is consistently high, while the model performs best for strong modification (modification types of \textit{extremity} or \textit{certainty}), decently for neutrals (\textit{neutrality} or no modification), and worst on weak modification (\textit{subtlety} or \textit{uncertainty}). This indicates that large, conspicuous features are easier for the model to capture compared to small or more ambiguous features.

In terms of subject/object properties, human performance is also consistently high. In contrast, model performance is significantly worse for subject ellipsis (\textit{inter-utterance subject}), while remaining high for object ellipsis and \textit{no object} cases.

We also hypothesize that a large portion of the relations can actually be satisfied \textit{without} considering the objects, e.g. by simply predicting very dark dots as the subjects when the relation is \textit{darker} or \textit{darkest}. To distinguish such easy cases, we consider a relation as \textit{ignorable object} if the relation can be satisfied even if we ignore the objects (i.e. remove all object relations) based on gold referents. Our result verifies that there are indeed many cases of \textit{ignorable object}, and they seem slightly easier for the model to satisfy.

\begin{table}[h!]
\centering \small
\def\arraystretch{1.0}
\newcommand{\intermidrule}{\cmidrule{2-13}}  
\setlength{\tabcolsep}{4pt}
\setlength{\aboverulesep}{0pt}
\setlength{\belowrulesep}{0pt}
\setlength{\extrarowheight}{.75ex}
\begin{tabular}{lccaca}
\toprule
\multicolumn{2}{c}{Models} & \multicolumn{2}{c}{\texttt{NUMREF}} & \multicolumn{2}{c}{Human} \\
\midrule
value & mod-type & diff. & \# valid & diff. & \# valid \\
\midrule
\multirow{3}{*}{xy-value} & strong & 86.06 & 39 & 89.15 & 37 \\
& neutral & 80.92 & 1,586 & 73.52 & 1,558\\
& weak & 80.35 & 200 & 53.53 & 198 \\
\midrule
\multirow{3}{*}{color} & strong & 66.23 & 15 & 91.80 & 15 \\
& neutral & 56.98 & 234 & 60.14 & 232\\
& weak & 37.73 & 68 & 28.55 & 66 \\
\midrule
\multirow{3}{*}{size} & strong & 3.60 & 8 & 4.29 & 8 \\
& neutral & 2.67 & 337 & 2.70 & 320\\
& weak & 1.95 & 105 & 1.58 & 104 \\
\bottomrule
\end{tabular}
\caption{
Absolute difference in comparative relations (number of valid predictions shown in shade).
}
\label{difference_comparison}
\end{table}

In Table \ref{difference_comparison}, we study the effect of modification based on the  \textit{absolute difference} between subject and object features in comparative relations. \footnote{\textit{Left/right} for x-value, \textit{above/below} for y-value, \textit{lighter/darker} for color and \textit{smaller/larger} for size.}

In human annotation, the absolute difference naturally increases as the modification gets stronger. While model predictions also show this tendency, their results seem less sensitive to modification (particularly for locational features, i.e. xy-value) and may not be reflecting their full effect.

\section{Discussion and Conclusion}
\label{sec:discussion_and_conclusion}

In this work, we focused on the recently proposed OneCommon Corpus as a suitable testbed for fine-grained language understanding in visually grounded dialogues. To analyze its linguistic structures, we proposed a novel framework of annotating spatial expressions in visual dialogues. We showed that our annotation can be conducted reliably and efficiently by leveraging referring expressions prevalent in visual dialogues, while capturing important linguistic structures such as PAS, modification and ellipsis. Although our current analysis is limited to this domain, we expect that upon appropriate definition of spatial expressions, argument roles and canonicalization, the general approach can be applied to a wider variety of domains: adapting and validating our approach in different domains (especially with more realistic visual contexts) are left as future work.

Secondly, we proposed a simple idea of incorporating \textit{numerical constraints} to improve exophoric reference resolution. We expect that a similar approach of identifying and incorporating semantic constraints (e.g. coreferences and spatial constraints) is a promising direction to improve the model's performance even further.

Finally, we demonstrated the advantages of our annotation for investigating the model's understanding of visually grounded dialogues. Our tests are completely agnostic to the models and only require referent predictions made by each model. By designing simple tests like ours (Subsubsection \ref{subsubsec:spatial_attributes}/\ref{subsubsec:spatial_relations}), we can diagnose the model's performance at the granularity of canonical attributes/relations under consideration: such analyses are easy to extend (by adding more tests) and critical for verifying what capabilities current models have (or do not have). Based on further analyses (Subsubsection \ref{subsubsec:further_analyses}), we also revealed various linguistic structures that affect model performance: we expect that capturing and studying such effects will be essential for advanced model probing in visual dialogue research.

Overall, we expect our framework and resource to be fundamental for conducting sophisticated linguistic analyses of visually grounded dialogues.

\section*{Acknowledgements}
This work was supported by JSPS KAKENHI Grant Number 18H03297 and NEDO SIP-2 ``Big-data and AI-enabled Cyberspace Technologies.” We also thank the anonymous reviewers for their valuable suggestions and comments.

\bibliographystyle{acl_natbib}
\bibliography{emnlp2020}

\appendix

\section{Annotation Examples and Details}
\label{sec:annotation_examples}

\begin{figure}[h!]
\centering
\includegraphics[width=0.96\columnwidth]{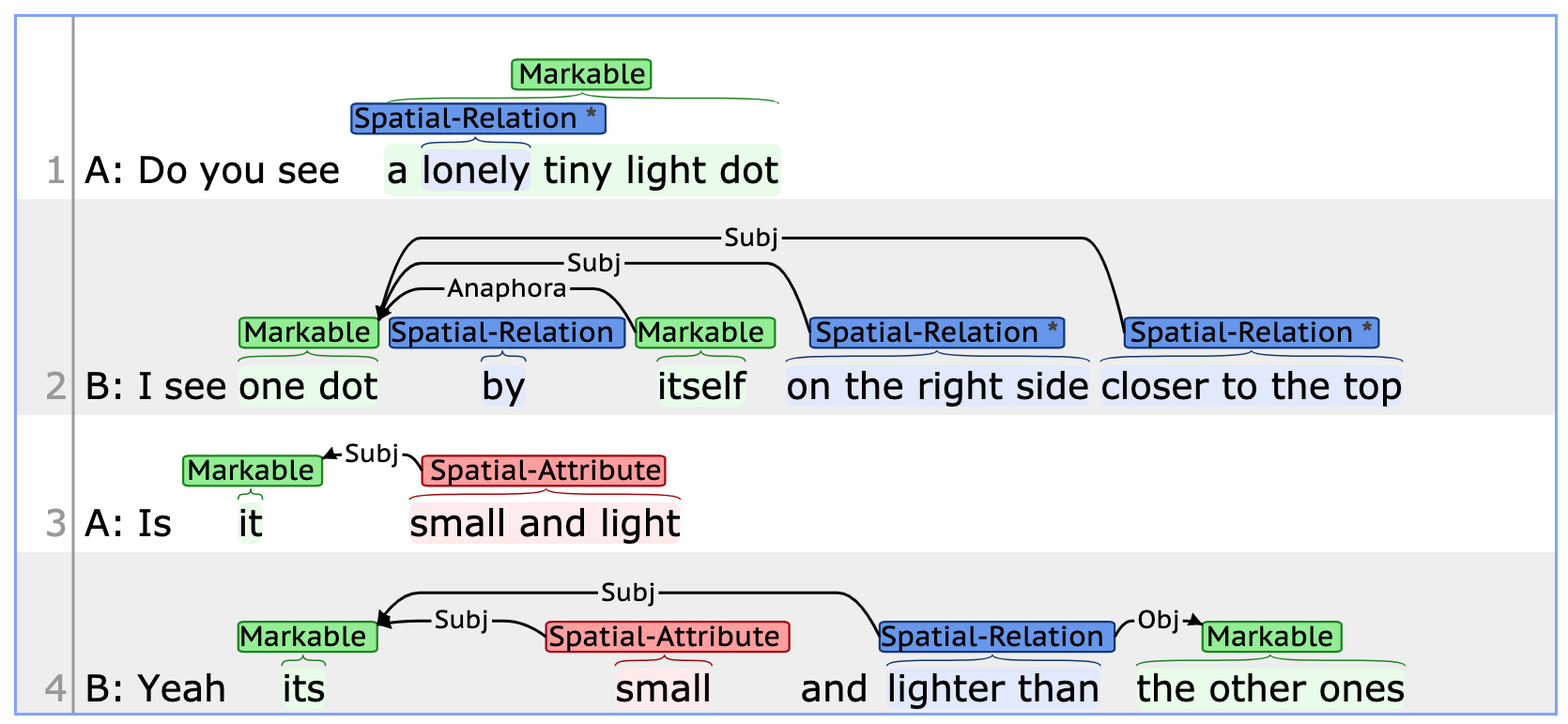}
\caption{Example with spatial attributes.}
\label{fig:spatial_attribute}
\end{figure}

Here, we show additional examples of our spatial expression annotation. In Figure \ref{fig:spatial_attribute}, we show an example dialogue annotated with \textit{spatial attributes} (colored in red). Since our goal is \textit{not} to achieve strict inter-annotator agreement but to conduct efficient and useful analysis, we allow certain flexibility in determining the spans of spatial expressions: for instance, the coordinated spatial expression (``\underline{small and light}'') can be annotated as a single expression or as different expressions (``\underline{small} and \underline{light}''). Copulas (\textit{is}, \textit{being}), articles (\textit{a}, \textit{the}), particles (\textit{to}, \textit{with}) and modifiers were allowed to be either omitted or included in spatial expressions. Spans were allowed to be non-contiguous, but must be annotated at the token level and restricted to be within a single utterance. Note that spatial attributes (\textit{tiny}, \textit{light}) in the first markable (``a lonely tiny light dot'') are not annotated, since they are inside the markable and their spans and subjects are relatively obvious.

In terms of argument identification, we prioritize markables in the following manner:

\begin{enumerate}
\item Markables in the present utterance (i.e. same utterance as the spatial expression).

\item Markables in the closest previous utterance of the \textit{same speaker}.

\item Markables in the closest previous utterance of \textit{different speakers}.
\end{enumerate}

As long as these priorities are satisfied, we did not distinguish between coreferences. Furthermore, for object identification, we did not distinguish between markables which include/exclude subject referents: for example, the object markable for \textit{lighter} in ``I have [three dots], [two] dark and [one] \underline{lighter}'' could be either \textit{three dots} or \textit{two}.

\begin{figure}[h!]
\centering
\includegraphics[width=0.96\columnwidth]{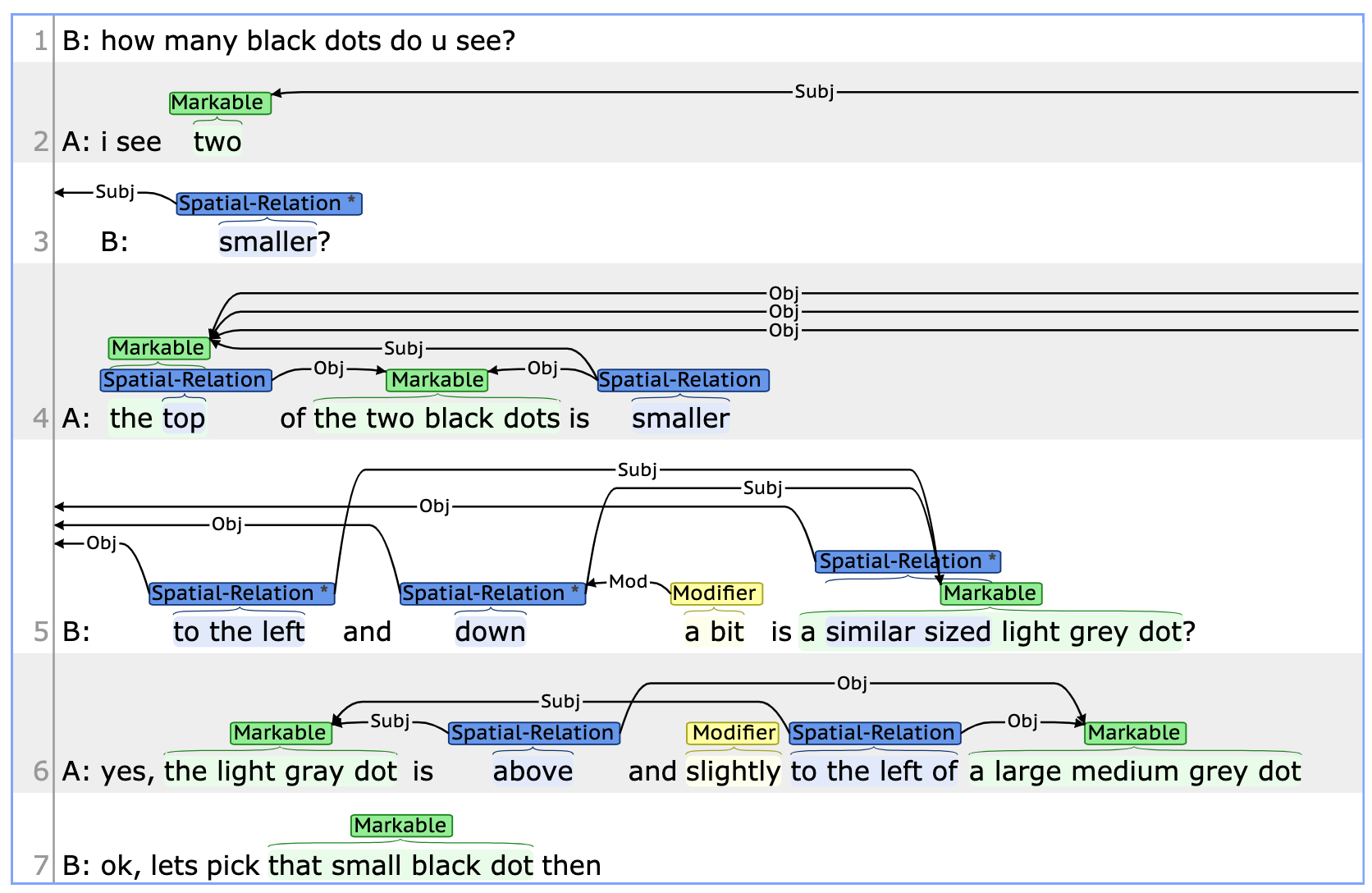}
\caption{Example with subject ellipsis.}
\label{fig:subject_ellipsis}
\end{figure}

In Figure \ref{fig:subject_ellipsis}, we show an example dialogue where the subject markable only appears in the previous utterance (``\underline{smaller}?'' in B's utterance), which demonstrates the case of \textit{subject ellipsis}. Note that since we only detect expressions that contain \textit{specific spatial information} of the visual context, we do not annotate \textit{black dots} in the first interrogative utterance (``how many black dots do u see?'').

\begin{figure}[h!]
\centering
\includegraphics[width=0.96\columnwidth]{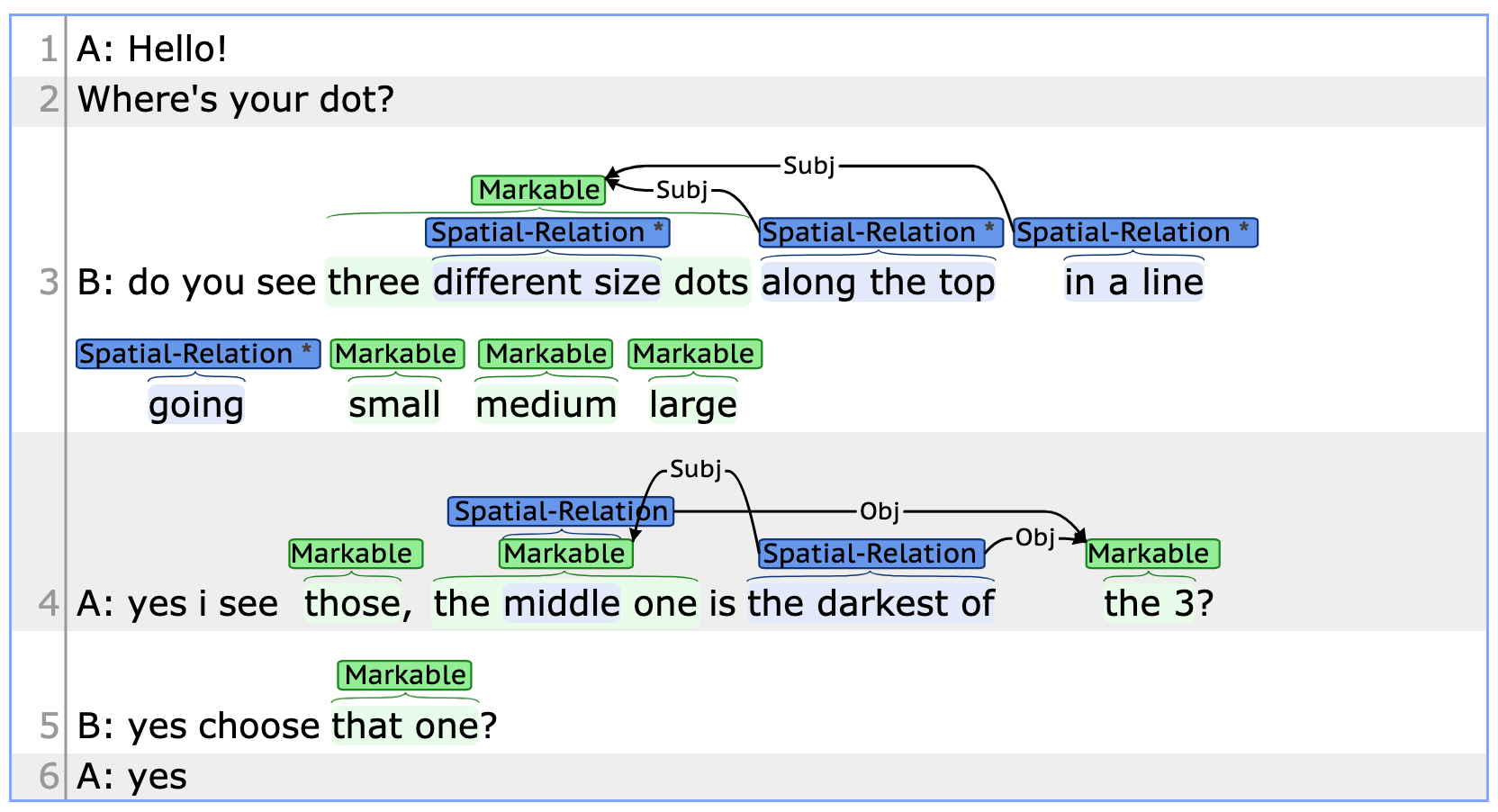}
\caption{Example with unannotatable relation.}
\label{fig:unannotatable}
\end{figure}

In Figure \ref{fig:unannotatable}, we show an example dialogue with \textit{unannotatable} relation (``\underline{going} [small], [medium], [large]'') which cannot be captured based on the simple argument roles of subjects and objects. In general, similar strategies of enumeration are difficult to be captured, as well as predications with \textit{exceptions} (such as ``[All dots] are \underline{dark} except [one dot]'') or cases with \textit{bundled} subjects (``[Two dots] are \underline{dark} and \underline{darker}'').

Finally, we only annotate \textit{explicit} spatial attributes and relations: therefore, we do not annotate \textit{implicit} relations such as \textit{darker} in ``One is dark and the other is light gray'', although it is inferable. When the spans are difficult to annotate, annotators were encouraged to make the best effort to capture the constructions which refer to specific spatial information.

\section{Annotation Results}
\label{sec:annotation_results}

\begin{table}[th]
\centering \small
\begin{tabular}{ccc}
\toprule
Annotation & \% Agreement & Cohen's $\kappa$ \\
 \midrule
Attribute Start & 98.5 & 0.84 \\
Relation Start & 95.1 & 0.77 \\
Modifier Start & 98.7 & 0.82 \\
\bottomrule
\end{tabular}
\caption{
Additional results of our reliability analysis.
}
\label{additional_reliability_results}
\end{table}

In Table \ref{additional_reliability_results}, we show the results of token level agreement for the \textit{starting positions} of spatial expressions and modifiers. Despite having certain freedom as discussed in Appendix \ref{sec:annotation_examples}, we can verify that these also have reasonably high agreement.

\begin{table}[th]
\centering \small
\begin{tabular}{lcc}
\toprule
 & Attribute & Relation \\
 \midrule
\% mod-subtlety & 1.06 & 8.12 \\
\% mod-extremity & 9.00 & 2.16 \\
\% mod-uncertainty & 7.41 & 4.26 \\
\% mod-certainty & 0.27 & 1.40 \\
\% mod-neutrality & 19.31 & 0.67 \\
\% mod-negation & 0.53 & 0.42 \\
\bottomrule
\end{tabular}
\caption{
Additional statistics of our spatial expression annotation.
}
\label{additional_annotation_statistics}
\end{table}

In Table \ref{additional_annotation_statistics}, we show the frequency of each modification types. Based on these results, we can see that \textit{neutrality} is the most common type of modification for spatial attributes (as in \textit{medium gray}, \textit{medium sized}), and \textit{subtlety} and \textit{uncertainty} to be the most common types for spatial relations. It is interesting to note that the frequencies of modification types vary significantly with spatial attributes and relations, except for \textit{negation}.

\begin{table*}[th]
\centering \small
\def\arraystretch{1.0}
\newcommand{\intermidrule}{\cmidrule{2-13}}  
\setlength{\tabcolsep}{4pt}
\setlength{\aboverulesep}{0pt}
\setlength{\belowrulesep}{0pt}
\setlength{\extrarowheight}{.75ex}
\begin{tabular}{lccc}
\toprule
Category & Relation & Unique & Examples \\
\midrule
\multirow{7}{*}{Direction}\phantom{..} & \textit{left} & 150 & to the left (78), on the left (35), left most (5), furthest left (2)  \\
 & \textit{right} & 192  & to the right (120), on the right (38), lower right (6), to the northeast (1) \\
 & \textit{above} & 190 & above (118), top (92), on top (33), up (17), higher (10), just above (4) \\
 & \textit{below} & 179 & below (88), bottom (56), lower (38), down (14), lowest (7), beneath (4) \\
 & \textit{horizontal} & 19 & horizontal (12), in a horizontal line (4), side by side (3), across from (1) \\
 & \textit{vertical} & 29 & vertical (7), on top of (5), on a vertical line (4), aligned vertically with (1) \\
 & \textit{diagonal} & 38 & diagonal (5), in a diagonal line (5), sloping down to the right (1), slanted (1) \\
\midrule
\multirow{3}{*}{Proximity} & \textit{near} & 59 & close together (63), cluster (32), next to (28), close to (22), near (13) \\
 & \textit{far} & 21 & far (5), away from (4), set apart from (1), a ways above (1), a distance from (1) \\
 & \textit{alone} & 13 & by (38), lonely (30), alone (21), lonesome (1), isolated (1) \\
\midrule
\multirow{2}{*}{Region} & \textit{interior} & 47 & middle (41), in the middle (19), between (9), in the center of (2) \\
 & \textit{exterior} & 46 & close to the border (5), all around (1), on the outside of (1), surrounding (1) \\
\midrule
\multirow{6}{*}{Color} & \textit{lighter} & 22 & lighter (102), lighter than (10), lighter gray (8), larger lighter (4) \\
 & \textit{lightest} & 11 & lightest (28), lightest shade (3), the lightest of (2), lightest and smallest (2) \\
 & \textit{darker} & 30 & darker (130), darker than (16), smaller and darker (4), darker in color (3) \\
 & \textit{darkest} & 10 & darkest (40), smallest darkest (2), the darkest of (1), darkest/largest of (1) \\
 & \textit{same} & 9 & same color (9), identical looking (2), similar shades (1), equally black (1) \\
 & \textit{different} & 11 & different shades (3), different sizes and shades (2), of varying shades (1) \\
\midrule
\multirow{6}{*}{Size} & \textit{smaller} & 17 & smaller (209), smaller than (5), smaller and lighter (4), tinier (1) \\
 & \textit{smallest} & 8 & smallest (40), tiniest (4), smallest darkest (2), smallest of (1) \\
 & \textit{larger} & 32 & larger (178), bigger than (7), larger in size (2), double the size of (1) \\
 & \textit{largest} & 10 & largest (41), biggest (11), largest of (2), biggest one of (1) \\
 & \textit{same} & 32 & same size (24), same sized (12), similar in size (5), identical in size (3) \\
 & \textit{different} & 8 & different sizes (3), of different sizes (1), varying sizes (1), opposite in sizes (1) \\
\bottomrule
\end{tabular}
\caption{
Unique numbers and examples of spatial relations which imply each canonical relation (frequencies shown in parentheses). 
}
\label{canonical_relation_examples}
\end{table*}

\begin{table*}[h!]
\centering \small
\def\arraystretch{1.0}
\newcommand{\intermidrule}{\cmidrule{2-13}}  
\setlength{\tabcolsep}{4pt}
\setlength{\aboverulesep}{0pt}
\setlength{\belowrulesep}{0pt}
\setlength{\extrarowheight}{.75ex}
\begin{tabular}{lcc}
\toprule
Modification & Unique & Examples \\
\midrule
Subtlety & 27 & slightly (235), a little (48), a bit (35), a tiny bit (8), very slightly (5) \\
Extremity & 15 & very (87), much (17), pretty (8), quite (3), really (2) \\
Uncertainty & 36 & almost (85), about (49), kind of (23), smallish (6), not completely (3)  \\
Certainty & 13 & directly (28), exactly (2), perfect (2), almost exactly (2)  \\
Neutrality & 16 & medium (59), med (9), fairly (4), mid-size (3), slightly medium (1)    \\
Negation & 4 & not (17), isn't (1), not perceptibly (1)  \\
\bottomrule
\end{tabular}
\caption{
Unique numbers and examples of modifiers with each modification type (frequencies in parentheses). 
}
\label{modification_type_examples}
\end{table*}

In Table \ref{canonical_relation_examples} and \ref{modification_type_examples}, we show the statistics and examples of canonical relations and modification types annotated for our analyses. Note that a single expression can imply multiple canonical relations (e.g. ``identical looking'' implies \textit{same color} and \textit{same size}) or no canonical relation at all (e.g. ``forms a triangle''). In contrast, a modifier can have only one modification type: for instance, \textit{almost exactly} is considered to have the overall modification type of \textit{certainty}.

\section{Experiment Setup}
\label{sec:experiment_setup}

We use the dataset, baselines, hyperparameters and evaluation metrics publicly available at \url{https://github.com/Alab-NII/onecommon}.

In order to collect model predictions for all dialogues and markables, we randomly split the whole dataset into 10 equal sized bins $z_i \: (i \! \in \! \{0,1,2,...,9\})$ and at each round $r \! \in \! \{0,1,2,...,9\}$ we use $z_{r \Mod{10}}$, $z_{r + 1 \Mod{10}}$, ..., $z_{r + 7 \Mod{10}}$ for model training, $z_{r + 8 \Mod{10}} $ for validation, and $z_{r + 9 \Mod{10}}$ for testing. We report the mean and standard deviation of the entity-level accuracy and markable-level exact match rate in these 10 rounds of the experiments.

In our \texttt{NUMREF} model, we train a separate module for predicting the number of referents based on a simple MLP (single layer, 256 hidden units). Reference resolution and number prediction are trained jointly with the loss weighted by 32:1. We conducted minimal hyperparameter tuning since the results did not change dramatically.

\section{Size Distribution Plots}
\label{sec:size_distribution_plots}

\begin{figure}[ht]
\centering
\includegraphics[width=0.97\columnwidth]{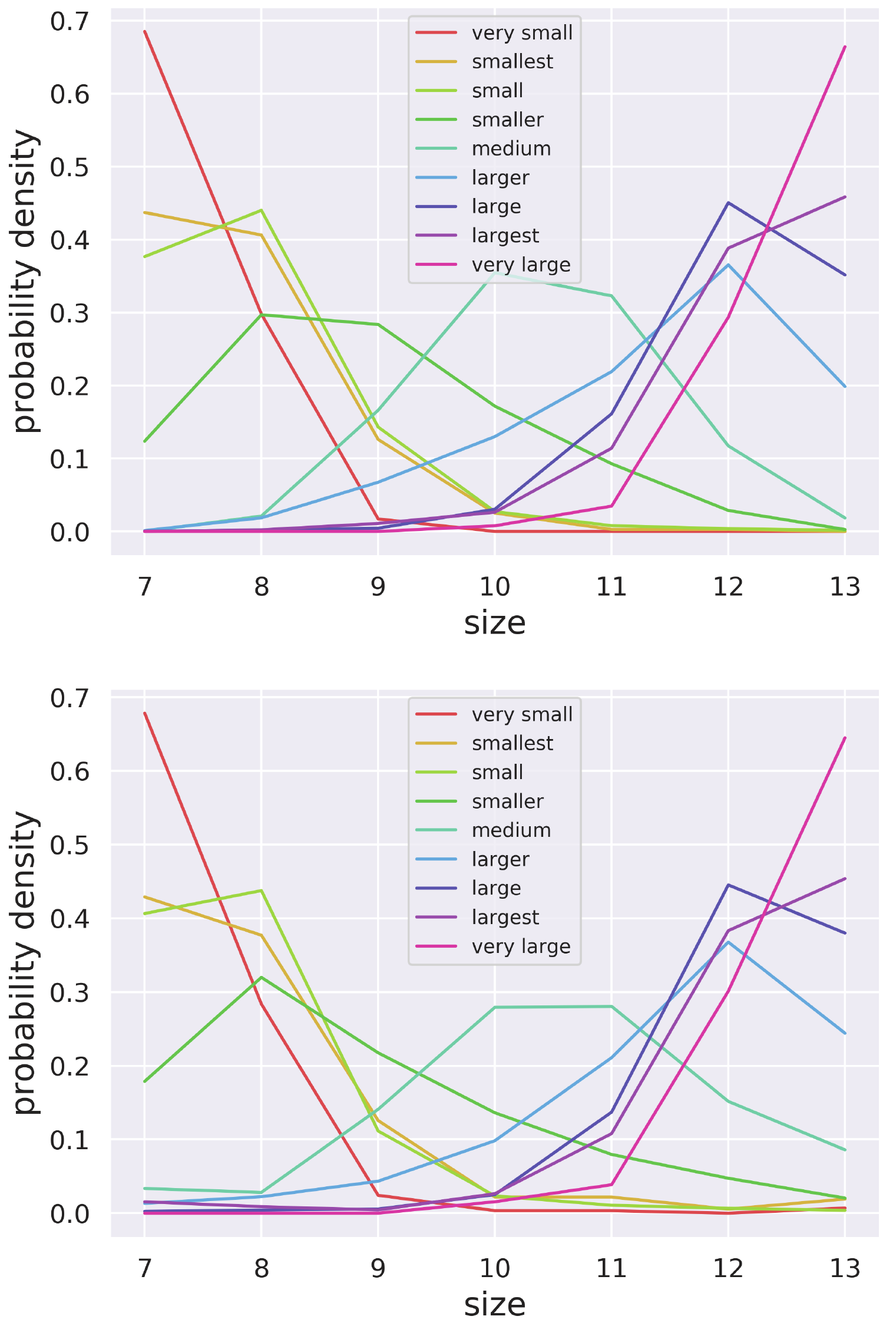}
\caption{Referent size distributions (top is human, bottom is \texttt{NUMREF}).}
\label{fig:referent_size}
\end{figure}

Figure \ref{fig:referent_size} shows the referent \textit{size} distributions based on human annotation (top) and \texttt{NUMREF} predictions (bottom). We can verify that the two distributions look almost identical for all common expressions, as observed in the color distributions.

\section{Canonical Relation Tests}
\label{sec:relation_test_algorithms}

For canonical relation tests, we only use relations that are \textit{not negated} and have all arguments in the \textit{same speaker}'s utterances (so that referent predictions are based on the same player's observation). As illustrative examples, we show the algorithms for testing the \textit{horizontal} relation (Algorithm \ref{alg:horizontal_test}), \textit{near} relation (Algorithm \ref{alg:near_test}), \textit{interior} relation (Algorithm \ref{alg:interior_test}) and \textit{same color} relation (Algorithm \ref{alg:same_color_test}). Note that each algorithm can take a variety of inputs, such as \textit{all referents} including both subjects and objects ($\mathcal{A}$) or \textit{all observable entities} of the player ($\mathcal{E}$).

\begin{algorithm}[ht]
\small
\DontPrintSemicolon
\SetAlgoNoEnd

\KwIn{all referents $\mathcal{A}$}
\KwOut{boolean $satisfy$, boolean $valid$}
$valid \leftarrow |\mathcal{A}| > 1$\\
\If{$valid$}{
	\tcp{Conduct linear regression and check if coeficient is small}
	$reg.fit(\mathcal{A})$\\
	$satisfy \leftarrow reg.coef < \frac{1}{3}$\\
}\Else{
	$satisfy \leftarrow $ False
}
\Return $satisfy$, $valid$
\caption{Test for \textit{horizontal} relation}
\label{alg:horizontal_test}
\end{algorithm}

\begin{algorithm}[ht]
\small
\DontPrintSemicolon
\SetAlgoNoEnd

\KwIn{all referents $\mathcal{A}$, observable entities $\mathcal{E}$}
\KwOut{boolean $satisfy$, boolean $valid$}
$valid \leftarrow |\mathcal{A}| > 1$ \\
\If{$valid$}{
	\tcp{Compute distance for every pair in the set}
	$A\_dists \leftarrow dist(x, y)$ \textbf{\upshape for} $x, y$ \textbf{\upshape in}  $combination(\mathcal{A})$\\
	$E\_dists \leftarrow dist(x, y)$ \textbf{\upshape for} $x, y$ \textbf{\upshape in}  $combination(\mathcal{E})$\\
	\tcp{Check if mean distance is smaller}
	$satisfy \leftarrow valid \, \wedge$ \\
	$\,\,\, mean(A\_dists) < mean(E\_dists)$
}\Else{
	$satisfy \leftarrow $ False
}
\Return $satisfy$, $valid$
\caption{Test for \textit{near} relation}
\label{alg:near_test}
\end{algorithm}

\begin{algorithm}[ht]
\small
\DontPrintSemicolon
\SetAlgoNoEnd
\KwIn{subject referents $\mathcal{S}$, object referents $\mathcal{O}$, boolean $no\_object$}
\KwOut{boolean $satisfy$, boolean $valid$}
\If{$no\_object$}{
	\tcp{If any subject referent is far from the center, satisfy is False}
	$valid \leftarrow |\mathcal{S}| > 0 $\\
	$satisfy \leftarrow valid$ \\
	$center \leftarrow (0, 0)$\\
	\For{$s \in \mathcal{S}$}{
		\If{$ dist(s, center) > 120$}{
	    	$satisfy \leftarrow $ False
	    }
	}
}\Else{
	\tcp{If any subject referent is outside the region of objects, satisfy is False}
	$valid \leftarrow |\mathcal{S}| > 0 \, \wedge \, |\mathcal{O}| > 1$\\
	$satisfy \leftarrow valid$ \\
	\For{$s \in \mathcal{S}$}{
		\If{$(s.x \! < \! min(\mathcal{O}.x) \vee max(\mathcal{O}.x) \! < \! s.x) \, \wedge$ \\
		$(s.y \! < \! min(\mathcal{O}.y) \vee max(\mathcal{O}.y) \! < \! s.y)$}{
	    	$satisfy \leftarrow $ False
	    }
	}
}

\Return $satisfy$, $valid$
\caption{Test for \textit{interior} relation}
\label{alg:interior_test}
\end{algorithm}

\begin{algorithm}[ht]
\small
\DontPrintSemicolon
\SetAlgoNoEnd

\KwIn{all referents $\mathcal{A}$}
\KwOut{boolean $satisfy$, boolean $valid$}
$valid \leftarrow |\mathcal{A}| > 1$\\
\tcp{Check if range of color is smaller than the threshold}
$satisfy \leftarrow valid \, \wedge$ \\
$\,\,\, max(\mathcal{A}.color) \! - \! min(\mathcal{A}.color) \! < \! 30$\\
\Return $satisfy$, $valid$
\caption{Test for \textit{same color} relation}
\label{alg:same_color_test}
\end{algorithm}

\end{document}